\documentclass[11pt,a4paper]{article}
\usepackage[hyperref]{naaclhlt2018}
\usepackage{times}
\usepackage{latexsym} 
\usepackage{url}
\usepackage{graphicx}

\usepackage{amsmath}
\usepackage{algorithm}
\usepackage[noend]{algpseudocode}
\usepackage[latin1]{inputenc}

\usepackage{lipsum}% http://ctan.org/pkg/lipsum

\aclfinalcopy % Uncomment this line for the final submission
\def\aclpaperid{1744} %  Enter the acl Paper ID here

\title{Accelerating NMT Batched Beam Decoding\\ with LMBR Posteriors for Deployment}

\author{Gonzalo Iglesias$^\dag$ \ \ \ \ \\\And
	William Tambellini$^\dag$ \ \ \  \\\And
	Adrià De Gispert$^{\dag\ddag}$ \\\And
	Eva Hasler$^\dag$\\\And
	Bill Byrne$^{\dag\ddag}$ \\ \\
	\hspace{-15cm} $^\dag$SDL Research \\
	\hspace{-15cm} \tt \{giglesias|wtambellini|agispert|ehasler|bbyrne\}@sdl.com \\ \\
    \hspace{-15cm} $^\ddag$Department of Engineering, University of Cambridge, U.K.}

\date{}

\begin{document}
\maketitle
\begin{abstract}
We describe a batched beam decoding algorithm for NMT with LMBR n-gram posteriors, showing that LMBR techniques still yield gains on top of the best recently reported results with Transformers. We also discuss acceleration strategies for deployment, and the effect of the beam size and batching on memory and speed.
 
%We have 6 pages for this paper: \url{http://naacl2018.org/industry.html}.
% I think our submission could fall into "Deployed" category.

%From the web page: "Submissions must clearly identify one of the following three areas they fall into:
%
%\textbf{Deployed}: Must describe deployment of a system that solves a non-trivial real-world problem. The focus should be on describing the problem, its significance, decisions and tradeoffs made when making design choices for the solution, deployment challenges, and lessons learned.
% 
%\textbf{Discovery}: Must include results obtained from NLP applications in real world scenarios that result in insights that are interesting and actionable. These discoveries should reveal promising directions in their application areas, leading to further system or societal enhancements. For example, an actionable discovery from an analysis of call center transcripts may reveal that certain language choices negatively impact customer experience, leading to better training of service representatives and improved customer experience.
%
%
%\textbf{Emerging}: Submissions do not have to describe deployed systems but must have clear applications to industry to distinguish them from NAACL research papers. They may also provide insight into issues and factors that affect the successful use and deployment of natural language processing. Papers that describe enabling infrastructure for large-scale deployment of natural language processing techniques also fall in this category." 
\end{abstract}

\section{Introduction}

The advent of Neural Machine Translation (NMT) has revolutionized the market. Objective improvements~\cite{Sutskever2014,Bahdanau2015,Sennrich2016b,Gehring2017,Vaswani2017} and a fair amount of neural hype have increased the pressure on companies offering Machine Translation services to shift as quickly as possible to this new paradigm.

Such a radical change entails non-trivial challenges for deployment; consumers certainly
look forward to better translation quality, but do not want to lose all the good features
that have been developed over the years along with SMT technology. With NMT, real time decoding is challenging without GPUs, and still an avenue for research~\cite{Devlin2017}. Great speeds have been reported by
~\citet{Junczys-Dowmunt2016} on GPUs, for which batching queries to the neural model is essential.
Disk usage and memory footprint of pure neural systems are certainly lower than that of SMT systems, but at the same time GPU memory is limited and high-end GPUs are expensive. 
%but then we are now also constrained to the memory available in the GPU, and high end GPUs are expensive.

Further to that, consumers still need the ability to constrain translations; in particular, brand-related information is often as important for companies as translation quality itself, and is currently under investigation~\cite{Chatterjee2017,Hokamp2017,Eva2018}. It is also well known that pure neural systems reach very high fluency, often sacrificing adequacy~\cite{Tu2017,Zhang2017,Koehn2017}, and have been reported to behave badly under noisy conditions~\cite{Belinkov2018}. \citet{Stahlberg2017} show an effective way to counter these problems by taking advantage of the higher adequacy inherent to SMT systems via Lattice Minimum Bayes Risk (LMBR) decoding~\cite{Tromble2008}. This makes the system more robust to pitfalls, such as over- and under-generation~\cite{Feng2016,Meng2016,Tu2016} which is important for commercial applications. 
% With this strategy, the system can also fall back to the best phrase-based translation without additional computation if the best NMT translation is deemed unsuitable.
%\footnote{TODO. mention neurobabble, maybe examples? Mention PBMT safety net?}

In this paper, we describe a batched beam decoding algorithm that uses NMT models with LMBR n-gram posterior probabilities~\cite{Stahlberg2017}. Batching in NMT beam decoding has been mentioned or assumed in the literature, e.g.~\cite{Devlin2017,Junczys-Dowmunt2016}, but to the best of our knowledge it has not been formally described, and there are interesting aspects for deployment worth taking into consideration. 

We also report on the effect of LMBR posteriors on state-of-the-art neural systems, for five translation tasks. Finally, we discuss how to prepare (LMBR-based) NMT systems for deployment, and how our batching algorithm performs in terms of memory and speed. 

\section{Neural Machine Translation and LMBR}
\label{sec:beam-batching}
Given a source sentence $\mathbf x$, a sequence-to-sequence NMT model scores a candidate translation sentence $\mathbf y = y_1^T$ with $T$ words as:
\begin{align}
P_{NMT}(y_1^T|\mathbf x) = \prod_{t=1}^T P_{NMT}(y_t|y_1^{t-1}, \mathbf x)
\end{align}

where $P_{NMT}(y_t|y_1^{t-1}, \mathbf x)$ uses a neural function $f_{NMT}(\cdot)$. 
%Multiple queries to a neural network are commonly batched together, as this takes advantage of the massive parallelization provided e.g. by GPUs. Given that for decoding, NMT models are typically used with a left-to-right beam decoder with a small beam of size $B$, the decoder naturally lends itself to batching. 
To account for batching $B$ neural queries together, our abstract function takes the form of $f_{NMT}(\mathbf S_{t-1}, \mathbf y_{t-1}, \mathbf A)$ where $\mathbf S_{t-1}$ is the previous batch state with $B$ state vectors in rows, $\mathbf y_{t-1}$ is a vector with the $B$ preceding generated target words, and $\mathbf A$ is a matrix with the annotations~\cite{Bahdanau2015} of a source sentence. The model has a vocabulary size $V$.

The implementation of this function is determined by the architecture of specific models. The most successful ones in the literature typically share in common an attention mechanism that determines which source word to focus on, informed by $\mathbf A$ and $\mathbf S_{t-1}$. \citet{Bahdanau2015} use recurrent layers to both compute $\mathbf A$ and the next target word $y_t$. \citet{Gehring2017} use convolutional layers instead, and \citet{Vaswani2017} prescind from GRU or LSTM layers, relying heavily on multi-layered attention mechanisms, stateful only on the translation side. Finally, this function can also represent an ensemble of neural models.

Lattice Minimum Bayes Risk decoding computes n-gram posterior probabilities from an \textit{evidence space} and uses them to score a \textit{hypothesis space}~\cite{Kumar2004,Tromble2008,Blackwood2010}. It improves single SMT systems, and also lends itself quite nicely to system combination~\cite{Sim2007,deGispert2009}. \citet{Stahlberg2017} have recently shown a way to use it with NMT decoding: a traditional SMT system is first used to create an evidence space $\varphi_e$, and the NMT space is then scored left-to-right with both the NMT model(s) and the n-gram posteriors gathered from $\varphi_e$. More formally:
%\vspace{-0.5cm}
\begin{equation}
\begin{split}
%\hat{y} = \arg\max_y \sum_{t=1}^T L_{y_{t-n}^{t-1}y_t}(\varphi_e) \\
\hat{y} = \arg\max_y \sum_{t=1}^T (\overbrace{\Theta_0 + \sum_{n=1}^4 \Theta_n P_{LMBR}(y_{t-n}^t| \varphi_e)}^{\mathbf L(y_{t-n}^{t-1},y_t)} \\
+ \lambda \log P_{NMT}(y_t|y_1^{t-1}, \mathbf{x}))
\end{split}
\label{eq:nmt-mbr-posteriors}
\end{equation}

%where $\varphi_e$ is the evidence space used to compute the n-gram posteriors of up to order 4.
% and $L_{hy}$ is defined, for n-grams of order $4$, as:
%\begin{equation}
%L_{hy}(\varphi_e) = \Theta_0 + \sum_{n=1}^4 \Theta_n P_{LMBR}(hy, \varphi_e)
%\end{equation}

For our purposes $\mathbf L$ is arranged as a matrix with each row uniquely associated to an n-gram history identified in $\varphi_e$: each row contains scores for any word $y$ in the NMT vocabulary.

$\mathbf L$ can be precomputed very efficiently, and stored in the GPU memory. The number of distinct n-gram histories is typically no more than $500$ for our phrase-based decoder producing $200$ hypotheses. Notice that such a matrix  only containing $P_{LMBR}$ contributions would be very sparse, but it turns into a dense matrix with the summation of $\Theta_0$. Both sparse and dense operations can be performed on the GPU. We have found it more efficient to compute first all the sparse operations on CPU, and then upload to the GPU memory and sum the constant $\Theta_0$ in GPU\footnote{
Ideally we would want to keep $\mathbf L$ as a sparse matrix and sum $\Theta_0$ on-the-fly. However this is not possible with ArrayFire 3.6.}.

\begin{algorithm*}[t!]
	\caption{Batch decoding with LMBR n-gram posteriors}\label{batch-decoding}
	\begin{algorithmic}[1]
		\Procedure{DecodeNMT}{x, $\mathbf L$}
		\State $T \gets \text{Maximum target hypothesis length}$
		\State $\mathbf{b,y,q} \text{ indices and scores, with } \mathbf b_0 \gets \mathbf 0, \mathbf y_0 \gets \mathbf 0, \mathbf q_0 \gets \mathbf 0$ %\Comment{Size $B \times 1$}
		%		\State $\mathbf b_0 \gets \mathbf 0$ %\Comment{Size $B \times 1$}
		%		\State $\mathbf y_0 \gets \mathbf 0$ %\Comment{Size $B \times 1$}
		\State $\mathbf A \gets \text{Annotations for source sentence x}$
		\State $\mathbf S_0 \gets \text{Initial decoder state}$ %\Comment{Size $B \times H$}
		
		\State $F = \{\}$ \Comment{ Set of EOS survivors}
		%	\State $\mathbf L \gets \Call{PrecomputePosteriorMatrix}{M}$
		\For{\texttt{$t$ = $1$ to $T$}}
		\State $\mathbf P_t, \mathbf S_t \gets f_{NMT}(\mathbf S_{t-1},\mathbf y_{t-1}, \mathbf A)$  %\Comment{model logprobs $B \times V$ and state $B \times H$}
		\State $\mathbf h \gets \text{$B$ histories identified through $\mathbf b$, $\mathbf y$ and $t$} $ 
		\State $\mathbf P_t \gets \mathbf P_t +  \Call{GetMatrixByRows}{\mathbf L, \mathbf h}$ \Comment{Add LMBR contributions}
		%	\State $\mathbf P_t \gets \Call{ApplyPosteriors}{\mathbf P_t, \mathbf L, \mathbf b, \mathbf y, k}$
		%	\State $\Call{Mask}{\mathbf P_t}$
		\State $\mathbf b_t,\mathbf y_t,\mathbf q_t  \gets \Call{TopB}{\mathbf P_t + \mathbf q_{t-1}'}$ %\Comment{Returns matrix indices and scores}
		\State $\mathbf S_t \gets \Call{GetMatrixByRows}{\mathbf S_t, \mathbf b_t}$ %\Comment{Realigns state with top B survivors}
		\For{\texttt{$j$ = $0$ to $B - 1$}}
		\If {$ y_{tj} = \textit{EOS}$}
		\State $F \gets F \cup (\{t,j, q_{tj}\})$  \Comment{Track indices and score}
		\State $q_{tj} \gets -\infty$ \Comment{Mask out to prevent hypothesis extension} 
		\EndIf
		
		\EndFor
		\EndFor
		\State \textbf{return} $\Call{getBestHypothesis}{F, \mathbf b, \mathbf y}$
		
		\EndProcedure
	\end{algorithmic}
\end{algorithm*}
\section{NMT batched beam decoding}
Algorithm~\ref{batch-decoding} describes NMT decoding with LMBR posteriors using beam size $B$ equal to the batch size. Lines $2$-$5$ initialize the decoder; the number of time steps $T$ is usually a heuristic function of the source length. $\mathbf q$ will keep track of the $B$ best scores per time step, $\mathbf b$ and $\mathbf y$ are indices. 
%All three can be seen as matrices with size $T \times B$.
% Notice that the decoder state has $B$ rows. Each corresponds to states for $B$ distinct live hypotheses that we will be expanding at each time step.

Lines $7$-$16$ are the core of the batch decoding procedure. At each time step $t$, given $\mathbf S_{t-1}$, $\mathbf y_{t-1}$ and $\mathbf A$, $f_{NMT}$ returns two matrices: $\mathbf P_t$, with size $B \times V$, contains log-probabilities for all possible candidates in the vocabulary  given $B$ live hypotheses.  $\mathbf S_t$ is the next batch state. Each row in $\mathbf S_t$ is the vector state that corresponds to any candidate in the same row of $\mathbf P_t$ (line 8).

Lines $9$, $10$ add the n-gram posterior scores. Given the indices in $\mathbf b$ and $\mathbf y$ it is straightforward to read the unique histories for the $B$ open hypotheses: the topology of the hypothesis space is that of a tree because an NMT state represents the entire live hypothesis from time step $0$. Note that $b_{tj} < B$ is the index to access the previous word in $\mathbf y_{t-1}$. In effect, indices in $\mathbf b$ function as backpointers, allowing to reconstruct not only n-grams per time step, but also complete hypotheses. As discussed for Equation~\ref{eq:nmt-mbr-posteriors}, these histories are associated to rows in our matrix $\mathbf L$. Function $\Call{GetMatrixByRows}{\cdot}$ simply creates a new matrix of size $B \times V$ by fetching those $B$ rows from $L$. This new matrix is summed to $\mathbf P_t$ (line $10$).

In line $11$, we get the indices and scores in $\mathbf P_t + \mathbf q_{t-1}'$ of the top B hypotheses. These best hypotheses could come from any row in $\mathbf P_t$. For example, all B best hypotheses could have been found in row 0. In that case, the new batch state to be used in the next time step should contain copies of row 0 in the other $B-1$ rows. This is achieved
again with $\Call{GetMatrixByRows}{\cdot}$ in line 12. 

Finally, lines $13$-$16$ identify whether there are any end-of-sentence (EOS) candidates; the corresponding indices and score are pushed into stack $F$ and these candidates are masked out (i.e. set to $-\infty$) to prevent further expansion. In line $17$, $\Call{GetBestHypothesis}{F}$ traces backwards the best hypothesis in $F$, again using indices in $\mathbf b$ and $\mathbf y$. Optionally, normalization by hypothesis length happens in this step. 

It is worth noting that:
\begin{enumerate}
	\item If we drop lines 9, 10 we have a pure left-to-right NMT batched beam decoder.
	\item Applying a constraint (e.g. for lattice rescoring or other user constraints) involves masking out scores in $\mathbf P_t$ before line 11.
%	\item The algorithm can be easily enhanced with early pruning techniques~\cite{GoogleBridgingTheGap, Delaney2006}. We need to track the best complete hypothesis score seen in decoding. 
%	The surviving candidates obtained in line $11$ are scanned and the candidates that. 
	\item Because the batch size is tied to the beam size, the memory footprint increases with the beam.
	\item Due to the beam being used for both EOS and non EOS candidates, it can be argued that this empoverishes the beam and it could be kept in addition to non EOS candidates (either by using a bigger beam, or keeping separately). Empirically we have found that this does not affect quality with real models.
	\item  The opposite, i.e. that EOS candidates never survive in the beam for $T$ time steps, can happen, although very infrequently. Several pragmatic backoff strategies can be applied in this situation: for example, running the decoder for additional time steps, or tracking all EOS candidates that did not survive in a separate stack and picking the best hypothesis from there. We chose the latter.
\end{enumerate}

\begin{table*}[t!]
	\begin{center}		
		\begin{tabular}{|r|c|c|c|c|c|}
			\cline{2-6} 
			\multicolumn{1}{c|}{} & \multicolumn{3}{c|}{\textbf{WMT17}} & \multicolumn{2}{c|}{\textbf{WAT}}\tabularnewline
			\cline{2-6}
			\multicolumn{1}{c|}{} & \textbf{ger-eng} & \textbf{eng-ger} & \textbf{chi-eng} & \textbf{eng-jpn} & \textbf{jpn-eng}\tabularnewline
			\hline 
			\textbf{PBMT} & 28.9 & 19.6 & 15.8 & 33.4 & 18.0 \tabularnewline
			\hline
			\textbf{FNMT} & 32.8 & 26.1 & 20.8 & 39.1 & 25.3 \tabularnewline
			\hline 
			\textbf{LNMT} & 33.7 & 26.6 & 22.0 & 40.4 & 26.1 \tabularnewline
			\hline 
			\textbf{TNMT} & 35.2 & 28.9 & 24.8 & 44.6 & 29.4 \tabularnewline
			\hline 
			\textbf{LTNMT} & 35.4 & 29.2 & 25.4& 44.9 & 30.2 \tabularnewline
			\hline 
			\textbf{Best submissions}  & 35.1 & 28.3  & 26.4 & 43.3 & 28.4 \tabularnewline
			\hline
		\end{tabular}
	\end{center}
	\caption{Quality assessment of our NMT systems with and without LMBR posteriors for GRU-based (FNMT, LNMT) and Transformer models (TNMT, LTNMT). Cased BLEU scores reported on $5$ translation tasks.The exact PBMT systems used to compute n-gram posteriors for LNMT and LTNMT systems are also reported. The last row shows scores for the best official submissions to each task. }
	\label{tab:eng-ger-wmt17}
\end{table*}

\subsection{Extension to Sentence batching}
\label{sec:sentence-batching}
In addition to batching all $B$ queries to the neural model needed to compute the next time step for one sentence, we can do \textit{sentence batching}: this is,  we translate $N$ sentences simultaneously, batching $B \times N$ queries per time step.

With small modifications, Algorithm~\ref{batch-decoding} can be easily extended to handle sentence batching.
If the number of sentences is $N$, 
\begin{enumerate}
	\item Instead of one set $F$ to store EOS candidates, we need $F_1 ... F_N$ sets.
	\item For every time step, $\mathbf b_t, \mathbf y_t$ and $\mathbf q_t$ need to be matrices instead of vectors, and minor changes are required in $\Call{TopB}{\cdot}$ to fetch the best candidates per sentence efficiently.
	\item $\mathbf P_t$ and $\mathbf S_t$ can remain as matrices, in which case the new batch size is simply $B\cdot N$.
	\item The heuristic function used to compute $T$ is typically sentence specific.
%	\item For early pruning, the best complete hypothesis score tracked during decoding is also sentence specific.
\end{enumerate}

\section{Experiments}
\subsection{Experimental Setup}
We report experiments on English-German, German-English and Chinese-English language pairs for the WMT17 task, and Japanese-English and English-Japanese for the WAT task.  For the German tasks we use \textit{news-test2013} as a development set, and \textit{news-test2017} as a test set; for Chinese-English, we use \textit{news-dev2017} as a development set, and \textit{news-test2017} as a test set. For Japanese tasks we use the ASPEC corpus~\citep{Nakazawa2016}.

%[KYTEA tokenizer. German compound splitting.]

We use all available data in each task for training. In addition, for German we use back-translation data~\cite{Sennrich2016}. All training data for neural models is preprocessed with the byte pair encoding technique described by~\citet{Sennrich2016b}. 
We use Blocks~\cite{Merrienboer2015} with Theano~\cite{Bastien-Theano-2012} to train attention-based single GRU layer models~\cite{Bahdanau2015}, henceforth called \textbf{FNMT}. The vocabulary size is $50$K. Transformer models~\cite{Vaswani2017}, called here \textbf{TNMT}, are trained using the Tensor2Tensor package\footnote{\url{https://github.com/tensorflow/tensor2tensor}} with a vocabulary size of $30$K. 

Our proprietary translation system is a modular homegrown tool that supports pure neural decoding (FNMT and TNMT) and with LMBR posteriors (henceforce called \textbf{LNMT} and \textbf{LTNMT} respectively), and flexibly uses other  components (phrase-based decoding, byte pair encoding, etcetera) to seamlessly deploy an end-to-end translation system. 

FNMT/LNMT systems use ensembles of 3 neural models unless specified otherwise; TNMT/LTNMT systems decode with 1 to 2 models, each averaging over the last 20 checkpoints.

The Phrase-based decoder (\textbf{PBMT}) uses standard features with one single 5-gram language model~\cite{Heafield2013}, and is tuned with standard MERT~\cite{Och2003}; n-gram posterior probabilities are computed on-the-fly over  rich translation lattices, with size bounded by the PBMT stack and distortion limits. The parameter $\lambda$ in Equation~\ref{eq:nmt-mbr-posteriors} is set as 0.5 divided by the number of models in the ensemble. Empirically we have found this to be a good setting in many tasks.

Unless noted otherwise, the beam size is set to 12 and the NMT beam decoder always batches queries to the neural model. The beam decoder relies on an early preview of ArrayFire 3.6~\cite{Yalamanchili2015}\footnote{\url{http://arrayfire.org}}, compiled with CUDA 8.0 libraries. For speed measurements, the decoder uses one single CPU thread.
For hardware, we use an Intel Xeon CPU E5-2640 at 2.60GHz. The GPU is a GeForce GTX 1080Ti. We report cased BLEU scores~\cite{Papineni2002}, strictly comparable to the official scores in each task\footnote{\url{http://matrix.statmt.org/} and \url{http://lotus.kuee.kyoto-u.ac.jp/WAT/evaluation/index.html}}.

\begin{figure*}[t!]
	\centering
	\includegraphics[width=0.9\linewidth]{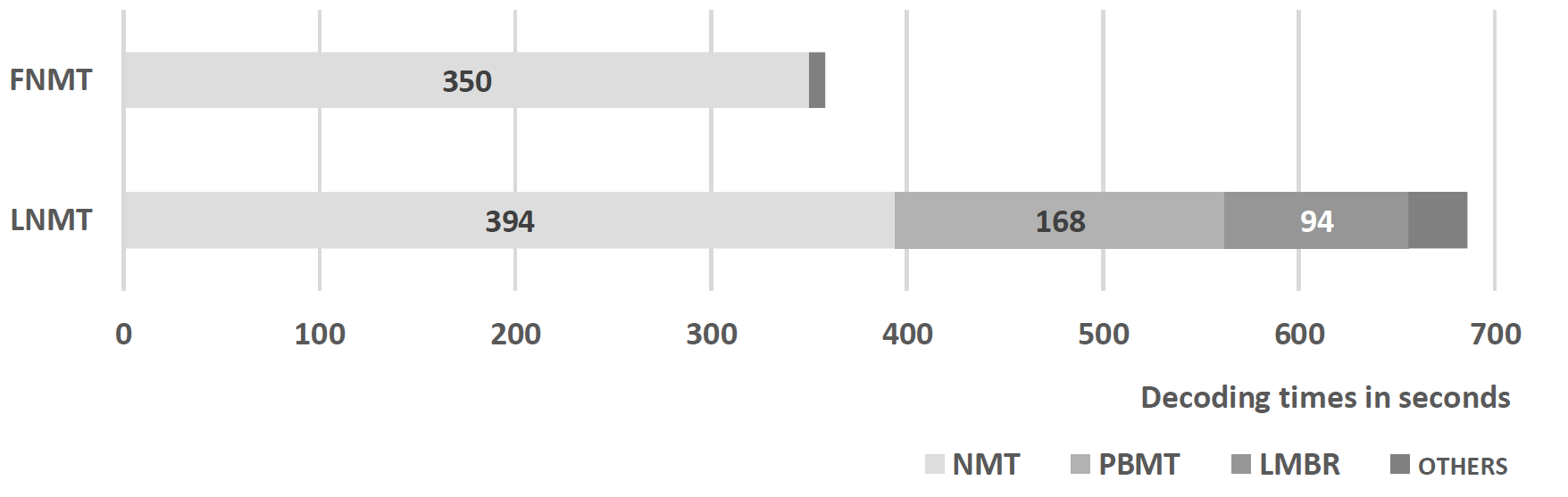}
	\caption{Accelerated FNMT and LNMT decoding times for newstest-2017 test set.}
	\label{fig:lnmt-fnmt-times2}
\end{figure*}

\subsection{The effect of LMBR n-gram posteriors}
Table~\ref{tab:eng-ger-wmt17} shows contrastive experiments for all five language pair/tasks. 
%These are research systems, i.e. there is no particular constraint on speed.
We make the following observations:
\begin{enumerate}
\item LMBR posteriors show consistent gains on top of the GRU model (LNMT vs FNMT rows), ranging from $+0.5$BLEU to $+1.2$BLEU. This is consistent with the findings reported by~\citet{Stahlberg2017}. 
\item The TNMT system boasts improvements across the board, ranging from $+1.5$BLEU in German-English to an impressive $+4.2$BLEU in English-Japanese WAT (TNMT vs LNMT). This is in line with findings by~\citet{Vaswani2017} and sets new very strong baselines to improve on.
\item Further, applying LMBR posteriors along with the Transformer model yields gains in all tasks (LTNMT vs TNMT), up to $+0.8$BLEU in Japanese-English. %LMBR techiques are yielding small gains even for English-German, where phrase-based lattices are $9$BLEU points below.
Interestingly, while we find that rescoring PBMT lattices~\cite{Stahlberg2016} with GRU models yields similar improvements to those reported by~\citet{Stahlberg2017}, we did not find gains when rescoring with the stronger TNMT models instead.
%This is in contrast to standard lattice rescoring techniques with TNMT, which interestingly did not yield improvements relative to the PBMT baseline. It is worth noting that rescoring with weaker GRU models does yield some improvements, in line with~\citet{Stahlberg2017}. 
\end{enumerate}

%[TODO: perhaps show examples of neurobabble for FNMT?]

\begin{figure*}[t!]
	\centering
	\includegraphics[width=0.9\linewidth]{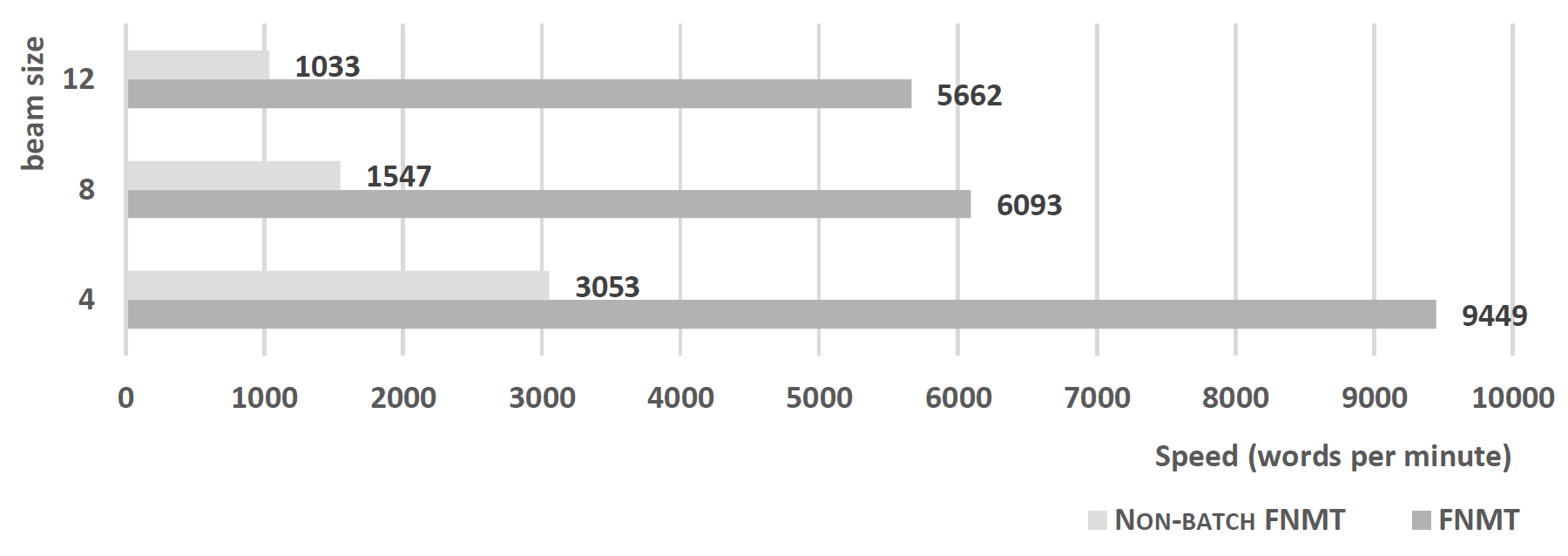}
	\caption{Batch beam decoder speed measured over newstest-2017 test set, using the accelerated FNMT system ($25.2$ BLEU for beam size = $4$).}
	\label{fig:fnmt-batch-vs-non-batch}
\end{figure*}
\begin{figure*}[t!]
	\centering
	\includegraphics[width=0.9\linewidth]{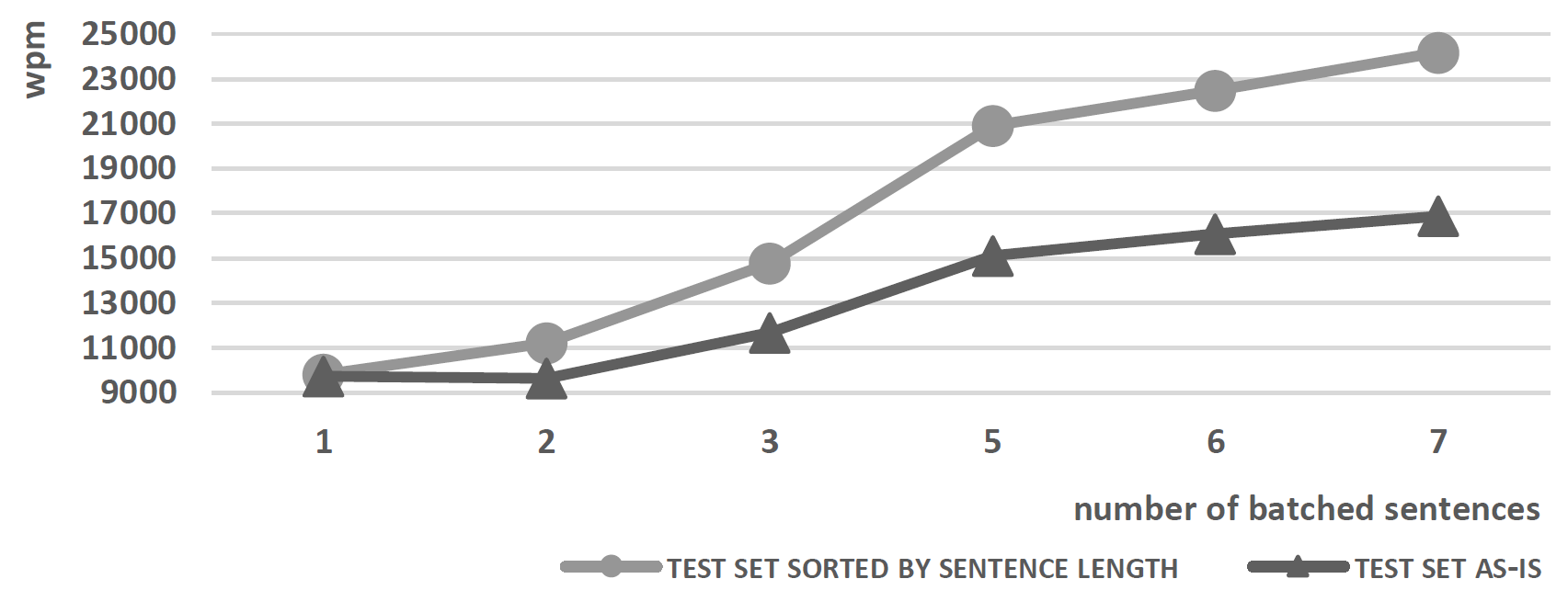}
	\caption{Batch beam decoder speed measured over newstest-2017 test set, using the accelerated eng-ger-wmt17 FNMT system ($26.1$ BLEU) with additional sentence batching, up to $7$ sentences.}
	\label{fig:sentence-batching}
\end{figure*}
\subsection{Accelerating FNMT and LNMT systems for deployment}

%\begin{figure}
%	\centering
%	\includegraphics[width=0.7\linewidth]{sentence-batching4}
%	\caption{}
%	\label{fig:sentence-batching4}
%\end{figure}

%Deployable NMT systems typically need to run at real-time speed with an acceptable memory footprint. In this section we discuss our current procedures to productize FNMT and LNMT systems\footnote{At the time of writing this paper, we are not deploying TNMT systems.}. Let us suppose that we wanted to deploy our English-German WMT17 LNMT system from Table~\ref{tab:eng-ger-wmt17}. 
There is no particular constraint on speed for the research systems reported in Table~\ref{tab:eng-ger-wmt17}. We now address the question of deploying NMT systems so that MT users get the best quality improvements at real-time speed and with acceptable memory footprint. As an example, we analyse in detail the English-German FNMT and LNMT case and discuss the main trade-offs if one wanted to accelerate them. Although the actual measurements vary across all our productised NMT engines, the trends are similar to the ones reported here.

In this particular case we specify a beam width of 0.01 for early pruning~\cite{GoogleBridgingTheGap, Delaney2006} and reduce the beam size to 4. 
%Reducing the beam size often does not result in substantial quality loss for NMT models~\cite{Britz2017}.
%Most of this degradation is due to reducing the beam size. The quality degradation varies in general for each translation system, but often does not result in substantial quality loss for NMT models~\cite{Britz2017}.
We also shrink the ensemble into one single big model\footnote{The file size of each $3$ individual models of the ensemble is 510MB; the size of the shrunken model is 1.2GB.} using the data-free shrinking method described by~\citet{Stahlberg2017b}, an inexpensive way to improve both speed and GPU memory footprint. 

In addition, for LNMT systems we tune phrase-based decoder parameters such as the distortion limit, the number of translations per source phrase and the stack limit. To compute n-gram posteriors we now only take a $200$-best from the phrase-based translation lattice. 

Table~\ref{tab:production} shows a contrast of our English-German WMT17 research systems versus the respective accelerated ones.

\begin{table}[h!]
\begin{center}
\begin{tabular}{|c|c|c||c|c|}
	\cline{2-5} 
	\multicolumn{1}{c|}{} & \multicolumn{2}{c||}{Research} & \multicolumn{2}{c|}{Accelerated}\tabularnewline
	\cline{2-5} 
	\multicolumn{1}{c|}{} & BLEU & speed & BLEU & speed\tabularnewline
	\hline 
	FNMT & 26.1 & 2207 & 25.2 & 9449\tabularnewline
	\hline 
	LNMT & 26.6 & 263 & 25.7 & 4927\tabularnewline
	\hline 
\end{tabular}
\end{center}
\caption{Cased BLEU scores for \textit{research} vs \textit{accelerated} English-to-German WMT17 systems. Speed reported in words per minute.}
\label{tab:production}
\end{table}

In the process, both accelerated systems have lost $0.9$ BLEU relative to the baseline. As an example, let us break down the effects of accelerating the LNMT system: using only $200$-best hypotheses from the phrase-based translation lattice reduces $0.3$ BLEU. Replacing the ensemble with a data-free shrunken model reduces another $0.2$ BLEU and decreasing the beam size reduces $0.4$ BLEU. The impact of reducing the beam size varies from system to system, although often does not result in substantial quality loss for NMT models~\cite{Britz2017}.

It is worth noting that these two systems share exactly the same neural model and parameter values. However, LNMT runs $4500$ words per minute (wpm) slower than FNMT. Figure~\ref{fig:lnmt-fnmt-times2} breaks down the decoding times for both the accelerated FNMT and LNMT systems. 
The LNMT pipeline also requires a phrase-based decoder and the extra component to compute the n-gram posterior probabilities. In effect, while both are remarkably fast by themselves (e.g. the phrase-based decoder is running at $20000$ wpm), these extra contributions explain most of the speed reduction for the accelerated LNMT system. In addition, the beam decoder itself is slightly slower for LNMT than for FNMT. This is mainly due to the computation of $\mathbf L$ as explained in Section~\ref{sec:beam-batching}. Finally, the respective GPU memory footprints for FNMT and LNMT are $4.1$ and $4.8$ GB.
% the GPU memory footprint is also higher for LNMT, partly due to $\mathbf L$ ($6$GB vs $4.1$GB).

\subsection{Batched beam decoding and beam size}
We next discuss the impact of using batch decoding and the beam size.
%effect of the beam size on the speed of our batch and non-batch FNMT  systems.
To this end we use the accelerated FNMT system ($25.2$ BLEU, $9449$ wpm) to decode with and without batching; we also widen the beam. Figure~\ref{fig:fnmt-batch-vs-non-batch} shows  the results.

The accelerated system itself with batched beam decoding and beam size of $4$ is $3$ times faster than without batching ($3053$ wpm). The GPU memory footprint is $1$ GB bigger when batching ($4.1$  vs $3.1$ GB). As can be expected, widening the beam decreases the speed of both decoders. The relative speed-up ratio favours the batch decoder for wider beams, i.e. it is 5 times faster for beam size 12. However, because the batch size is tied to the beam size, this comes at a cost in GPU memory footprint~(under $8$ GB).

\subsection{Sentence batching}
As described in Section~\ref{sec:sentence-batching}, it is straightforward to extend beam batching to sentence batching. Figure~\ref{fig:sentence-batching} shows the effect of sentence batching up to 7 sentences on our accelerated FNMT system.

Whilst the speed-up of our implementation is sub-linear, when batching 5 sentences the decoder runs at almost 21000 wpm, and goes beyond 24000 for 7 sentences. Thus, our implementation of sentence batching is $2.5$ times faster on top of beam batching. Again, this comes at a cost: the GPU memory footprint increases as we batch more and more sentences together, up to $11$ GB for $7$ sentences, which approaches the limit of GPU memory. 
 
Note that sentence batching does not change translation quality. For example, when translating 7 sentences, we are effectively batching 28 neural queries per time step. Indeed, each individual sentence is still being translated with a beam size of 4.

Figure~\ref{fig:sentence-batching} also shows the effect of sorting the test set by sentence length. Because sentences have similar lengths, less padding is required and hence we have less wasteful GPU computation. With $7$ batched sentences the decoder would run at barely $17000$ wpm, this is, $7000$ wpm less due to not sorting by sentence length. A similar strategy is common for neural training~\cite{Sutskever2014,Morishita2017}.

%These results are encouraging, and we are working to push this feature into production.

%Sentence batching should be particularly well suited e.g. for document translation or translation servers with a high incoming throughput. 
%Combining sentence batching with SMT technologies (e.g. phrase-based decoder, LMBR) for production can be somewhat challenging because typically older technologies have not been designed to support sentence batching.

\section{Conclusions}
We have described a left-to-right batched beam NMT decoding algorithm that is transparent to the neural model and can be combined with LMBR n-gram posteriors. Our quality assessment with Transformer models~\cite{Vaswani2017} has shown that LMBR posteriors can still improve such a strong baseline in terms of BLEU. Finally, we have also discussed our acceleration strategy for deployment and the effect of batching and the beam size on memory and speed.

%\section*{Acknowledgments}
\bibliography{naaclhlt2018}
\bibliographystyle{acl_natbib}

%\appendix

\end{document}